# Effects of Images with Different Levels of Familiarity on EEG

Ali Saeedi and Ehsan Arbabi

*School of Electrical and Computer Engineering, College of Engineering, University of Tehran, Iran*

*{ali.saeedi, earbabi}@ut.ac.ir*

**Abstract:** Evaluating human brain potentials during watching different images can be used for memory evaluation, information retrieving, guilty-innocent identification and examining the brain response. In this study, the effects of watching images, with different levels of familiarity, on subjects' Electroencephalogram (EEG) have been studied. Three different groups of images with three familiarity levels of "unfamiliar", "familiar" and "very familiar" have been considered for this study. EEG signals of 21 subjects (14 men) were recorded. After signal acquisition, pre-processing, including noise and artifact removal, were performed on epochs of data. Features, including spatial-statistical, wavelet, frequency and harmonic parameters, and also correlation between recording channels, were extracted from the data. Then, we evaluated the efficiency of the extracted features by using p-value and also an orthogonal feature selection method (combination of Gram-Schmitt method and Fisher discriminant ratio) for feature dimensional reduction. As the final step of feature selection, we used 'add-r take-away l' method for choosing the most discriminative features. For data classification, including all two-class and three-class cases, we applied Support Vector Machine (SVM) on the extracted features. The correct classification rates (CCR) for "unfamiliar-familiar", "unfamiliar-very familiar" and "familiar-very familiar" cases were 85.6%, 92.6%, and 70.6%, respectively. The best results of classifications were obtained in pre-frontal and frontal regions of brain. Also, wavelet, frequency and harmonic features were among the most discriminative features. Finally, in three-class case, the best CCR was 86.8%.

## 1 Introduction

Studying human brain potentials, during watching different images, has recently attracted researchers' attention. Such studies are usually conducted for memory evaluation [1, 2], coding and retrieving information in the brain [3], guilty-innocent identification [4-6], and examining the brain response during observation of object-face images or meaningful/familiar-meaningless/unfamiliar images [7-14]. The images used in such studies are usually divided into two groups of face and non-face images and each is categorized into some subgroups. For example, in the face study, some studies have conducted on the differences between brain signals when watching face-object images and familiar-unfamiliar face images [7-11]. In the area of non-face images, the differences of brain signals during watching meaningful and meaningless images and the effect of images repetitions have mostly studied [12-15]. In most studies, two methods of functional imaging (functional Magnetic Resonance Imaging (fMRI)) and brain potentials recording (Electroencephalogram (EEG) and Event-Related Potentials (ERP)) are used. Each one has their own advantages and disadvantages. Though expensive, functional imaging has very high spatial resolution. On the other hand, the methods like EEG signal acquisition have higher time resolution and less cost than functional methods [4].

The representation and formation of a mental image from a stimulus requires the synchronization of a high number of neurons. According to research studies, the neural activities for a stimulus displayed for the first time to users, are very higher than a state in which the same stimulus is displayed for next times [12, 15, 16]. Gruber and Müller [15] studied the effects of picture repetition on ERP.

Meaningful images [17] were used in this scenario. Each image was consecutively shown to each user for maximally three times. The research results revealed that showing the image for the first time leads to an amplitude increase in the gamma band of EEG. This increase was less in the second and third repetitions than in the first time. In other words, neural activities and consequently EEG amplitude are reduced comparing to the first display. This is called the neural saving phenomenon stating that the brain uses its previous information to identify a non-new image.

In another research [14], Gruber and Müller studied the effect of repeating both meaningful and meaningless images. For meaningful images, they again used the image bank in [17] and for meaningless images they used random relocation of meaningful image pixels. In their study, there was no consecutive image repetition and at least one image was shown between showing two similar images. Repeating meaningful images was followed by a general reduction in the spectral power of brain signal, while this was inversed for meaningless images and an increase was observed in this case. These results are totally consistent with neural saving phenomenon, because meaningful images are recognizable for the brain in subsequent repetitions and the brain does not need to be involved in a high activity. For meaningless images, however, as they are not recognizable for the person even in the subsequent repetitions, the neural activities are still high.

Seeck et al. [9] studied the potentials relating to the way of categorizing the objects and face images in brain. They used images of flowers, faces, portraits, and tools as stimuli. As results showed, the face-flower images had the lowest and the images of face-portrait and flower-tool images had the highest correlations. This shows the different behaviors of the brain in categorizing objects and faces. Results also indicated that for faces -and not for portraits- there was a N170 component in the ERP. Other studies also specify this component to faces [8, 10, 11]. This has been confirmed in Seeck's studies, as the N170 was not found when the images of other body organs such as eyes, hands and fingers were displayed.

Goffaux et al. [10] examined the effects of face and non-face images in three cases of ordinary, low and high spatial resolution. They used one face image and an image of an automobile in the three above-mentioned states (totally 6 images). Based on the results, it was concluded that the perception components relating to face recognition are possibly in low spatial frequencies.

In references in which the image of objects was used as a stimulus, two types of images (meaningful and meaningless) were applied. Meaningful images are those that the individual knows its quiddity and, on the other hand, meaningless images are those that he/she knows that they are not the images of a particular object (e.g. a spiral or an imaginative design). In this study, which is originally based on a work done in 2014 in the School of Electrical and Computer Engineering at University of Tehran, we examined the effects of meaningful images with three levels of familiarity, including "unfamiliar", "familiar", and "very familiar". In our definition, unfamiliar images are those that people do not know about their quiddity and application, while they are meaningful for them. For example, a medical device with specific application is an unfamiliar image which may not be recognized by ordinary individuals. Individuals may know that it is a device any way, but they do not know what it exactly is and what its application is. Familiar images are those that people are completely aware of their quiddity and application, but they do not deal with them in their everyday life. Finally, very familiar images are those that people not only know their quiddity and application, they also see them every day. For example, the image of a yellow cab (taxi) is expected to be a very familiar image for people who lives in some cities such as Tehran (capital of Iran).

The main difference of this research with previous ones is in the type of images used as stimuli. In previous studies, at least a set of images were meaningless. Here, all three sets are meaningful images that participants: 1- have never seen and do not know, 2- have never/rarely seen but know, and 3- have frequently seen and know. As all objects in our environment are meaningful (even if we do not know their quiddity and application), this research can help to identify the brain function when dealing with images with different levels of familiarities. In addition, the results of this study can be used in applications in which knowledge about familiarity level of people with devices/images is desired.

## 2 Methods and Materials

### 2.1 Signal Acquisition

Twenty-one graduate students of University of Tehran (14 males, mean age: 24.6 ± 2.35 y) participated in this study. All of them were right-handed and had normal or corrected-to-normal visual acuity. EEG signals were recorded by NrSign 3840 EEG Device in Medical Equipment Laboratory of the School of Electrical and Computer Engineering at University of Tehran. Signal recording was done according to 10-20 EEG system, using 21 Ag electrodes. The forehead electrode was used as the reference electrode. Also, in order to control eye movement and blinking artifacts, electrooculogram (EOG) was recorded. The sampling frequency was 500 Hz.

Subjects sat on a comfortable chair at the distance of 45 cm from the display monitor. Unfamiliar, familiar, and very familiar concepts were then explained to them. To be assured that subjects well understood these concepts, they were asked to clarify them. For each of these concepts, subjects were shown two sample images. These images were excluded from the main images. The participants were asked to look at (concentrate on) the images from their familiarity perspective.

We used eight images for each of three categories (total: 24). Very familiar images included computer mouse, computer keyboard, the university entrance gate, the university card reader, the university self-service plate, the university classrooms, etc. Familiar images included igloo, crossbow, quad decker bus, battle axe, hot air balloon, etc. Unfamiliar images included different unknown and unseen devices (such as special laboratory instruments). Some images were chosen by searching in internet and some others were prepared by the authors. Some parts of images (e.g. backgrounds) were first removed and the images were then manipulated in order to be in the same size and brightness. The selected images for the main scenario were pre-verified by asking other people, different from the main participating subjects, about their familiarity level. In other words, it was tried to keep subjects' opinions about the familiarity level of the images as similar as possible. A different image (image of a bird) was used to keep participants' attention high. Subjects had to press a key when seeing this image.

The images were displayed in a random order and each image was shown 20 times (totally $20 \times 24 = 480$ display). Each image was appeared for 2 seconds in a white background on a computer monitor. After that the monitor showed a null image (white background) for another 2 seconds. The scenario was divided into two parts to prevent participants' tiredness. During the test, subjects had to look at the images while avoiding any conversation or movement. After the first part of the test and before starting the second part, in order to get informed about the subjects' opinion about the familiarity level of each image, a questionnaire was filled by them.

## 2.2 Data Processing

### 2.2.1 Pre-Processing

After applying 0.5 – 35 Hz band-pass filter to the recorded signals, the most important challenge is to eliminate artifacts from EEG. Independent component analysis (ICA) was used to eliminate artifacts such as blinking and eye movement. We also considered the signal recorded from 200 ms to 2000 ms after displaying each image as a single EEG epoch. Thus, the time length of each EEG epoch corresponding to each image-display was 1800 ms.

### 2.2.2 Feature Extraction

Extracting proper features plays an effective role in the quality of categorization. A proper feature is a feature that can appropriately differentiate all classes. We used common and classic features which will be independently explained in the following sections. General information about the extracted features can be found in Table 1.

Table 1. Number of features in each feature set.

| Feature type | Number of features |
|---|---|
| Statistical-Time | 22 |
| Frequency | 16 |
| Harmonic Parameters | 15 |
| Wavelet coefficient moments | 16 |
| Correlation between channels | 18 |

**Statistical-time features**

Statistical-time features are extracted in time-domain. In this paper, we used skewness, kurtosis and also Hjorth parameters including activity, mobility, and complexity as statistical-time features [18]. Each parameter is defined as follows:

Skewness, defined according to Eq. (1), shows the asymmetry of a distribution in a way that the skewness of a Gaussian distribution is zero. Distributions with longer left tails and longer right tails have negative and positives values of skewness, respectively. In Eq. (1, $m_2$ and $m_3$ are the second and the third central moments in time domain. The $k^{th}$ central moment is obtained by Eq. (2.

$$\text{Skewness} = \frac{m_3}{m_2\sqrt{m_2}} \tag{1}$$

$$m_k = \frac{1}{n}\sum_{i=1}^{n}(x(i)-\bar{x})^k \tag{2}$$

Kurtosis, defined according to Eq. (3, shows the peak or flatness of the target distribution relative to the normal distribution. In Eq. (3, $m_2$ and $m_4$ are the second and the forth central moments in time domain (see Eq. (2).

$$\text{Kurtosis} = \frac{m_4}{(m_2)^2} \tag{3}$$

Activity is in fact the variance of signal in time domain which is a measure for determining the signal variations (relative to its average). It is defined as Eq. (4 for a one-dimensional signal in which $x(i)$, $\bar{x}$ and N are signal, signal's time average, and signal length, respectively.

$$\text{Activity} = \text{var}(x) = \frac{1}{N}\sum_{i=1}^{N}(x(i)-\bar{x})^2 \quad (4)$$

Mobility of a time signal ($x$) is defined according to Eq. (5, in which $x'$ is the first time-derivative of $x$.

$$\text{Mobility} = \frac{\sigma_{x'}}{\sigma_x} = \sqrt{\frac{\text{var}(x')}{\text{var}(x)}} = \frac{\text{Activity}(x')}{\text{Activity}(x)} \quad (5)$$

The last Hjorth parameter is complexity, which is defined as ratio of mobility of $x'$ to mobility of $x$. As Eq. (5) shows, complexity is related to the variance of $x''$ (i.e. the second time-derivative of $x$).

**Frequency features**

The frequency features include mode frequency, median frequency, and mean frequency [5], the relative spectral power (RSP) of frequency sub-bands, slow-wave index (SWI) [18], and harmonic parameters [19]. The mode frequency is the frequency with maximum power density (see Eq. (6)) and the mean frequency is the frequency for which the spectral power density is divided into two parts with equal energy (see Eq. (7)). The mean frequency can also be calculated by using Eq. (8.

$$\text{mode frequency} = \left\{f \mid S(f) = \max_f(S(f))\right\} \quad (6)$$

$$\text{median frequency} = f_{median} = \left\{f \mid \int_0^f S(f)df = \int_f^{+\infty} S(f)df\right\} \quad (7)$$

$$\text{mean frequency} = f_{mean} = \frac{\int_0^\infty f \times S(f)df}{\int_0^\infty S(f)df} \quad (8)$$

Different frequency bands are defined for brain signals. These bands include delta, theta, alpha, and beta. In some applications, they are divided into some sub-bands too. In this study, four frequency bands and 10 sub-bands have been considered (see Table 2) [18, 20].

Table 2. Frequency bands and sub-bands [18].

| Frequency band | Sub-band | Frequency range (Hz) |
|---|---|---|
| Delta | Delta 1 | (0.5-2) |
|  | Delta 2 | (2-4) |
| Theta | Theta 1 | (4-6) |
|  | Theta 2 | (6-8) |
| Alpha | Alpha 1 | (8-10) |
|  | Alpha 2 | (10-12) |
|  | Alpha 3 | (12-14) |
| Beta | Beta 1 | (14-16) |
|  | Beta 2 | (16-25) |
|  | Beta 3 | (25-35) |

The SWI is defined for delta, theta and alpha bands. Delta-slow-wave index (DSI), theta-slow-wave index (TDI), and alpha-slow-wave index (ADI) are defined in Eqs. (9 to (11, respectively. In these equations, BSP$_{sub\text{-}band}$ represents the spectral power of the signal in the target *sub-band* [18, 21].

$$\text{DSI} = \text{BSP}_{\text{delta}}/(\text{BSP}_{\text{theta}} + \text{BSP}_{\text{alpha}}) \tag{9}$$

$$\text{TSI} = \text{BSP}_{\text{theta}}/(\text{BSP}_{\text{delta}} + \text{BSP}_{\text{alpha}}) \tag{10}$$

$$\text{ASI} = \text{BSP}_{\text{alpha}}/(\text{BSP}_{\text{delta}} + \text{BSP}_{\text{theta}}) \tag{11}$$

The harmonic parameters include central frequency ($f_c$), band width ($f_\sigma$), and the spectral power amplitude in the central frequency ($S_{f_c}$), calculated by Eqs. 12 to 14. $P_{xx}$, $f_L$ and $f_H$ are estimated spectral power density, lower limit of frequency-band and higher limit of frequency-band, respectively.

$$f_c = \sum_{f_L}^{f_H} f P_{xx}(f) \bigg/ \sum_{f_L}^{f_H} P_{xx}(f) \tag{12}$$

$$f_\sigma = \left(\sum_{f_L}^{f_H} (f - f_c)^2 P_{xx}(f) \bigg/ \sum_{f_L}^{f_H} P_{xx}(f)\right)^{1/2} \tag{13}$$

$$S_{f_c} = P_{xx}(f_c) \tag{14}$$

**Wavelet transform coefficients**

Wavelet transform is a multi-resolution analysis that shows the similarity of the main signal and the shifted and scaled versions of a mother wavelet function. In other words, if the main signal is similar to the mother wavelet function in a particular scale, the wavelet transform coefficient in that scale will be greater than coefficients in other scales [4]. In practice, high and low-pass filters are employed to implement wavelet transform and the main signal is passed through these two filters for a number of stages, depending on a desired resolution. The output of the high-pass filter including high frequency information is called "details" and the output of the low-pass filter including low frequency information is called "approximation". In each stage, the output of high-pass filters remains unchanged and is stored. The output of low-pass filters is passed again through another high and a low-pass filter.

As the sampling frequency in this study was 500 Hz, signal should be passed six times in order to have aforementioned frequencies bands (Table 2). In this case, the outputs of the last four filters contains signals with frequency contents of (0-4), (4-8), (8-16), and (16-32) Hz. Here, instead of using coefficients, $1^{st}$, $2^{nd}$, $3^{rd}$ and $4^{th}$ moments of each of these four groups of coefficients are used as features. This was done for two reasons: first, there are a large number of points in each band which increase the size of feature vector. Second, the use of moments instead of main coefficients increases the robustness of features while keeping them discriminant. Therefore, 16 wavelet-features, i.e. 4 bands × 4 moments for each band, have been extracted.

**Correlation between channels**

For each epoch, time-correlation between channels (two by two) was considered as another group of features.

*2.2.3 Selecting Effective Features*

Feature selection is one of the most important stages in classification problems. Redundant features just occupy a large space and increase the calculation complexity. To find the best features, there should be a criterion by which the features should be examined and then the best of them is selected. This criterion is either classifier-independent or classifier-dependent. In literature, the former is called filter and the latter is called wrapper. Here, to find the best features independent from the classification method, we used two filter methods based on p-value criterion and Fisher criterion and one wrapper method based on "plus-r take-away-1" searching strategy [22].

**P-value criterion**

P-value is a criterion to assess how randomly a phenomenon occurs. If the estimated p-value is smaller than a defined significant level, the occurrence of the phenomenon is considered non-random. For feature selection, the phenomenon to be tested is the ability of a feature for classification. In this study, we used student's t-test to estimate p-value and find out whether a feature has significantly different values in two different classes or not. For each feature, if the difference is significant (p-value < 0.01), that feature is assumed to be appropriate. Among all appropriate features, 500 features with the lowest p-value were selected for all possible two-class cases.

**Fisher criterion**

Fisher criterion is another filter method that measures how good a feature can differentiate a number of classes. In two-class classification, the Fisher criterion is called Fisher discriminant ratio (FDR) and is defined by Eq. (15). Before measuring the FDR for 500 features obtained in previous stage, features were orthogonalized by the Gram–Schmidt orthogonalization algorithm. Gram–Schmidt orthogonalization algorithm increases the inter-class differentiability [22]. Applying FDR, 100 features were selected as the best features among the 500 orthogonalized features.

$$\text{FDR} = \frac{(m_1 - m_2)^2}{(\sigma_1^2 + \sigma_2^2)^2} \tag{15}$$

**Plus-r take-away-1 searching strategy**

As the last stage of feature selection, the number of features was reduced from 100 to 20 using "plus-r take-away-1" searching method – one of the alternative methods to compensate the slowness of floating forward selection method (we used prtools 4.1 [23]). Plus-r take-away-1 searching method is classifier-dependent; therefore, Support Vector Machine (SVM) with Gaussian kernel is used as classifier. To calculate the best value for sigma in each stage, a high step search was done. Having found the appropriate range for sigma, smaller steps were taken then. The best sigma values were between 0.8 and 0.9.

## 3   Results and Discussion

### 3.1   Two-Class Categorization

By applying SVM classifier on 20 selected features, Correct Classification Rate (CCR) for all two-class cases, including "unfamiliar-familiar", "unfamiliar-very familiar" and "familiar-very familiar", were found. Figure 1 presents CCR for each individual participant and Table 3 presents the mean and standard deviation of CCR for all participants in all two-class cases.

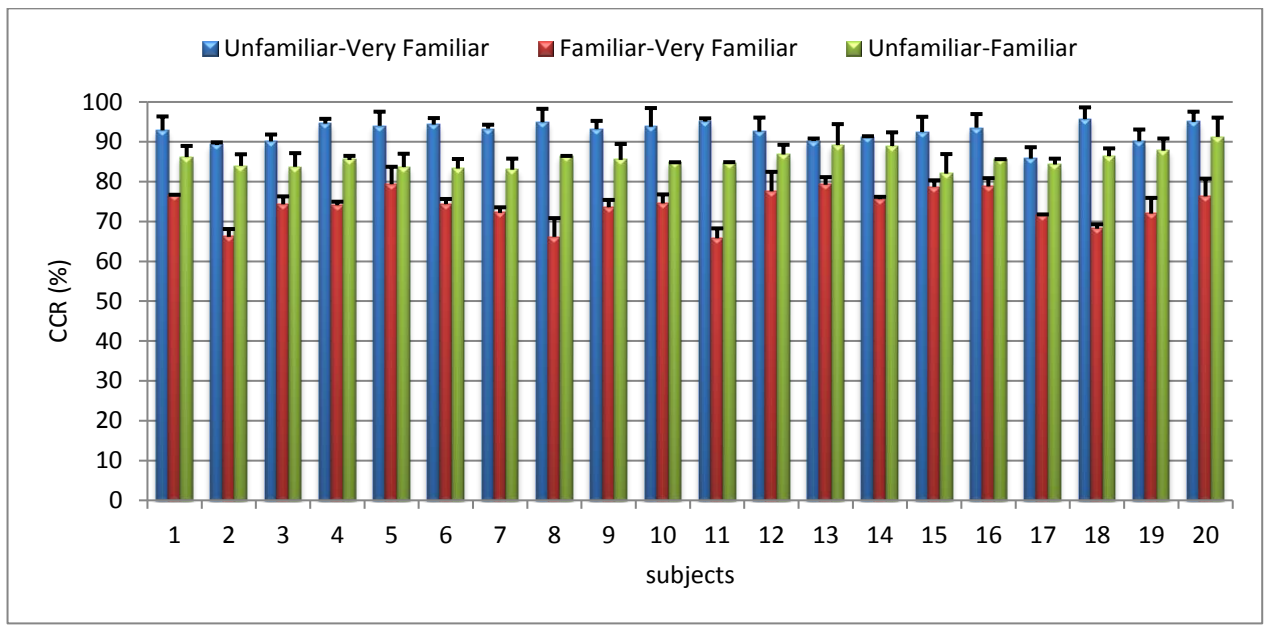

**Figure 1. CCR for all subjects in all two-class cases.**

High value of CCR for "unfamiliar-very familiar" images is normal, because the difference of the familiarity of these two groups is higher than others. In fact, the subject has never seen the unfamiliar images, but he/she frequently sees the very familiar images. A similar justification can also be applied to "unfamiliar-familiar" images except that the familiarity of familiar images is less than very familiar images. As mentioned in the introduction, people know the meaning and the concept of the familiar images, but they have much less confrontation with such images comparing with very familiar images. According to Figure 1, the lowest value of CCR was for "familiar-very familiar". Although the differences of familiar and very familiar images were explained to subjects, the CCR for "familiar-very familiar" images showed that their effects on EEG were not easily recognizable (comparing to other cases). In fact, the average CCR for "familiar-very familiar" was low because both groups contain meaningful images for subjects. In other words, subjects might unconsciously pay more attention to the quiddity of images rather that their familiarity. That is why in "unfamiliar-familiar" and "unfamiliar-very familiar" the average CCR was higher than "familiar-very familiar" –in which participants knew the quiddity of both groups. Remember that, although unfamiliar images were meaningful, participants did not know them and their application.

**Table 3. Mean and standard deviation of CCR in all two-class cases.**

|  | Unfamiliar-Very familiar | Familiar-Very familiar | Unfamiliar-Familiar |
|---|---|---|---|
| Mean of CCR (%) | 92.60 | 70.57 | 85.64 |
| Standard Deviation of CCR (%) | 2.35 | 2.57 | 2.16 |

The average appearance of each group of features (Table 1) among the best 20 selected features is also studies, considering all participants and all possible two-class cases. Figure 2 shows the average appearance of each group. It can be seen that the wavelet features have the best performance among the 20 selected features. In other words, in average, half of the selected features are wavelet features. After wavelet features, harmonic parameters and frequency features have the largest share. Channels correlation and statistical-time features had the lowest share.

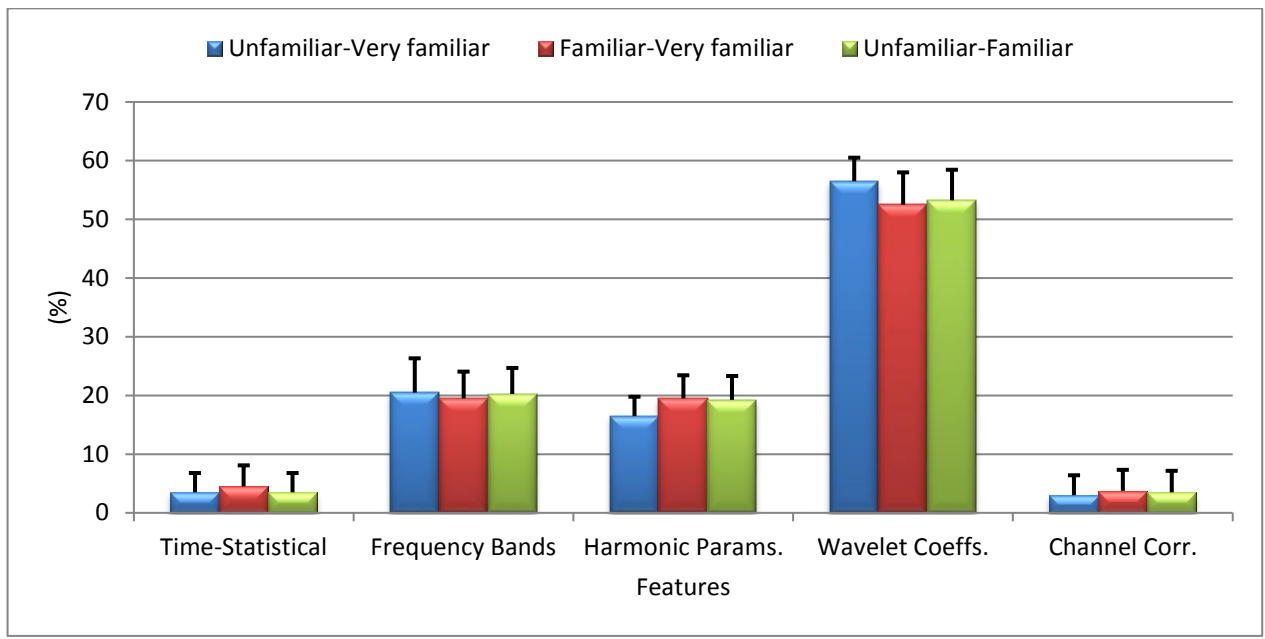

**Figure 2. Distribution of the 20 best features in 5 feature groups.**

To compare different regions of the brain, channels in which EEG were recorded were categorized into 6 regions. These regions include pre-frontal (Fp), frontal (F), central (C), temporal (T), parietal (P), and occipital (O). Table 4 presents the name and number of available electrodes in each region. The number of electrodes is important because the high performance of an area may be due to the higher number of electrodes in that specific region. Thus, regions that have the highest share in the 20 features are stated once without and once with normalizing to the number of electrodes in each region.

**Table 4. Name and number of electrodes used for signal acquisition.**

| Region name | Region electrodes | Number of electrodes |
|---|---|---|
| pre-Frontal | Fp1, Fp2 | 2 |
| Frontal | Fz, F8, F7, F4, F3 | 5 |
| Central | Cz, C4, C3 | 3 |
| Temporal | T6, T5, T4, T3 | 4 |
| Parietal | Pz, P4, P3 | 3 |
| Occipital | O1, O2 | 2 |

Figure 3 and Figure 4 show the un-normalized and normalized distribution of different brain regions among the best selected features, respectively. In un-normalized case, frontal and temporal regions have the highest share and the occipital area has the lowest share in the selected features. In normalized case, pre-frontal region has better performance than other regions.

### 3.2 Three-Class Categorization

Using results obtained by two-class classifications, we studied three-class classification. Accordingly, three-class classification was carried out by using the best 20 features obtained in two-class classifications (3 times). In three-class classification, we employed the SVM classifiers - with the same conditions explained in two-classification cases - and one-against-one method. The CCR for each participant and then the average CCR for all participants were calculated using the best 20 features of each two-class case. The results are presented in Table 5. According to Table 5, the best features which can recognize the effects of unfamiliar images on EEG in two-class cases (i.e. "unfamiliar-very familiar" and "unfamiliar-familiar" classifications) are also suitable for three-class case.

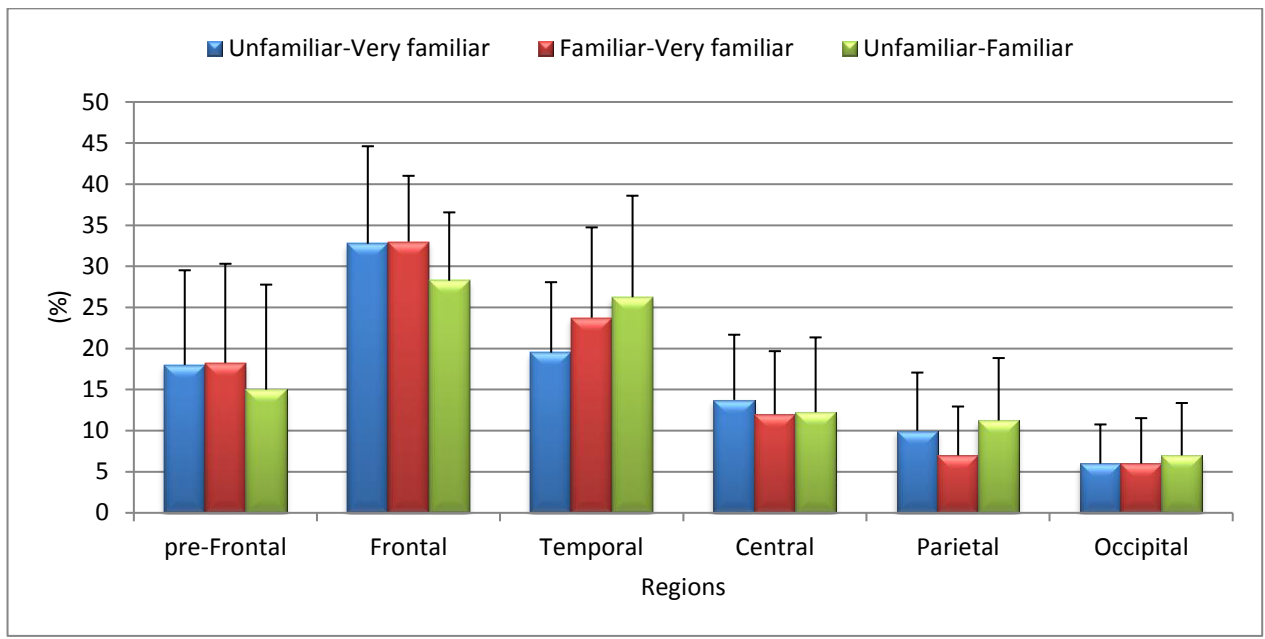

**Figure 3. Un-normalized distribution of the 20 best features in 6 brain regions.**

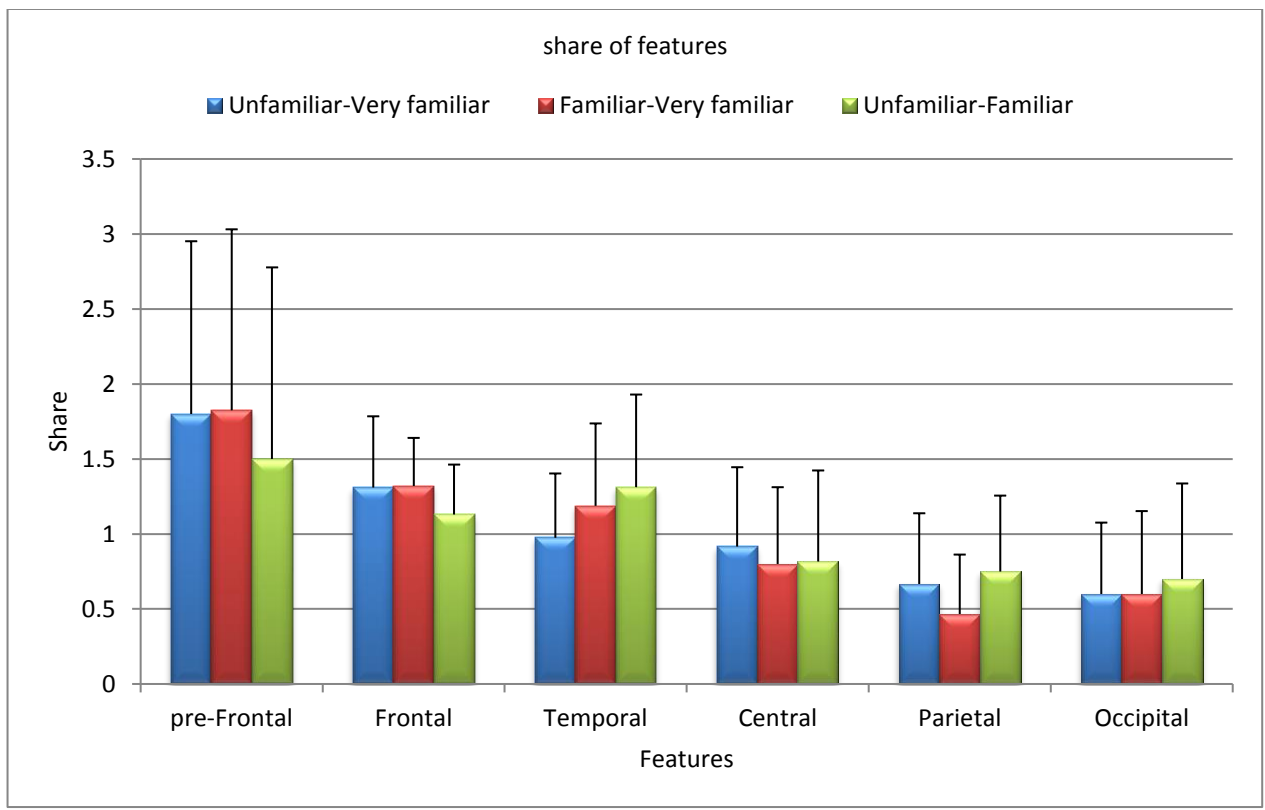

**Figure 4. Normalized distribution of the 20 best features in 6 brain regions.**

**Table 5. Comparison of the average CCR in two-class and three-class classification. Note that the CCRs of three-class cases are obtained using the corresponding features of two-class cases.**

|  | **Unfamiliar-Very Familiar** | | **Familiar-Very Familiar** | | **Unfamiliar-Familiar** | |
|---|---|---|---|---|---|---|
| Number of classes in classification | 3 | 2 | 3 | 2 | 3 | 2 |
| Average CCR (%) | 86.69 | 92.60 | 73.28 | 70.57 | 86.82 | 85.64 |

## 4 Conclusion

In this paper, we studied the effects of showing images with different levels of familiarity on subjects' EEG. The major difference between this study and the previous studies is using meaningful images and considering three groups of "unfamiliar", "familiar", and "very familiar" images. Unfamiliar images are those meaningful images that we are sure that subjects do not know and have not seen them before (e.g. the image of a laboratory device which is not publicly seen). Familiar images, on the other hand, are those images people know them but have never/rarely seen them (e.g. the image of a strange automobile). Very familiar images are those that people know and frequently deal with them (e.g. a yellow cab for a person lives in the city of Tehran).

By extracting different types of features, applying different feature selection methods and using SVM classifier, high CCR was achieved for recognizing the effects of unfamiliar images on EEG (in respect to the effects of familiar and very familiar images on EEG). However, lower classification rate was achieved when separating the effects of "familiar" images from the effects of "very familiar" images on EEG. The reason can be that familiar and very familiar images can be put in one group in terms of quiddity. Put it differently, subjects know both familiar and very familiar images and put them in the group of "I know what it is". They, however, do not know unfamiliar images and put them in the group of "I do not know what it is". Therefore, the classification of familiarity level turns unconsciously into a problem of "knowing or not knowing the quiddity". In applications aiming to evaluate memory, we can just use the group of "unfamiliar-very familiar" images.

The results show that wavelet and frequency features can contain more information for recognizing the effects of un/very/familiar images on EEG, comparing to time and statistical features. Wavelet transform features contain information both in time and frequency domains. Thus, it is natural that they outperform frequency features containing information in only frequency domain. In addition, the results show that frontal, pre-frontal and temporal regions of brain seem to be more active when subjects are concentrating on familiarity levels of different Images.

As a future work, one can use images of personal stuffs such as shoes, cell-phone, laptop, etc. as very familiar images in order to compare the effects of familiar images and very familiar images on EEG, more precisely. Another suggestion for improving this study can be changing the scenario of displaying images. Displaying two groups of images in separate scenarios instead of showing all three image groups together, may help to better evaluate the effects of images on EEG. In addition to EEG, event-related potential (ERP) analysis can be added to have a better and more precise classifications.

## Acknowledgements

The authors would like to thank Dr. Mohammad Ali Akhaee for his scientific consultation. In addition, the authors are thankful to Ms. Afsoon Khodaee, Ms. Mina Mirjalili and all who participated in EEG recording stage.